\documentclass[letterpaper]{article} 
\usepackage{aaai23}  
\usepackage{times}  
\usepackage{helvet}  
\usepackage{courier}  
\usepackage[hyphens]{url}  
\usepackage{graphicx} 
\usepackage{natbib}  
\usepackage{caption} 
\usepackage{soul}
\usepackage{url}
\usepackage[utf8]{inputenc}
\usepackage{xcolor}
\usepackage{amsfonts}
\usepackage{amsmath}
\usepackage{amsthm}
\usepackage{booktabs}
\usepackage{algorithm}
\usepackage{algorithmic}
\usepackage{setspace}
\usepackage{tablefootnote}
\usepackage{multirow}
\usepackage{makecell}
\usepackage[normalem]{ulem}
\usepackage{diagbox}
\usepackage{changes}

\usepackage{newfloat}
\usepackage{listings}
\DeclareCaptionStyle{ruled}{labelfont=normalfont,labelsep=colon,strut=off} 
\floatstyle{ruled}
\newfloat{listing}{tb}{lst}{}
\floatname{listing}{Listing}
\title{For the Underrepresented in Gender Bias Research: Chinese Name Gender Prediction with Heterogeneous Graph Attention Network}
\author {
    Zihao Pan\equalcontrib,
    Kai Peng\equalcontrib,
    Shuai Ling,
    Haipeng Zhang \thanks{*Corresponding author.}
}
\affiliations {
    ShanghaiTech University\\
    \{panzh, pengkai, lingshuai, zhanghp\}@shanghaitech.edu.cn
}

\begin{document}

\maketitle

\begin{abstract}

Achieving gender equality is an important pillar for humankind’s sustainable future. Pioneering data-driven gender bias research is based on large-scale public records such as scientific papers, patents, and company registrations, covering female researchers, inventors and entrepreneurs, and so on. Since gender information is often missing in relevant datasets, studies rely on tools to infer genders from names. However, available open-sourced Chinese gender-guessing tools are not yet suitable for scientific purposes, which may be partially responsible for female Chinese being underrepresented in mainstream gender bias research and affect their universality. Specifically, these tools focus on character-level information while overlooking the fact that the combinations of Chinese characters in multi-character names, as well as the components and pronunciations of characters, convey important messages. As a first effort, we design a Chinese Heterogeneous Graph Attention (CHGAT) model to capture the heterogeneity in component relationships and incorporate the pronunciations of characters. Our model largely surpasses current tools and also outperforms the state-of-the-art algorithm. Last but not least, the most popular Chinese name-gender dataset is single-character based with far less female coverage from an unreliable source, naturally hindering relevant studies. We open-source a more balanced multi-character dataset from an official source together with our code, hoping to help future research promoting gender equality.
\end{abstract}

\section{Introduction}
Recently, there have been increasing gender-equality studies, regarding female researchers~\cite{lariviere2013bibliometrics,huang2020historical}, inventors~\cite{jensen2018gender,koning2021we}, entrepreneurs~\cite{ritter2019man,van2021matching}, and STEM students~\cite{cimpian2020understanding}, based on large-scale public records such as scientific papers, patents, and company registrations. Given the fact that many of these datasets do not contain gender information, genders are usually inferred from individual names. Surprisingly, Chinese females are underrepresented in such research, though there is abundant comparable data in Chinese. One possible reason is the lack of reliable Chinese name-gender guessing tools that suit the standard of scientific purposes, compared with their English counterparts.

Ngender\footnote{https://github.com/observerss/ngender} is a basic tool for Chinese name-gender guessing, based on Naïve Bayes. It calculates the probability of a first name, often consisting of one or two Chinese characters, being a female name, by multiplying such probabilities of individual characters. Though straightforward, Ngender has two limitations. One is associated with Naïve Bayes itself -- it does not work for out-of-sample characters that the classifier has never seen.
Furthermore, it overlooks the knowledge from the combination of characters as well as the character components. The Ngender training data also deepens both limitations -- it only contains the numbers of times each available character appears in names of females and names of males, instead of the frequencies of complete first names under both genders.
Existing studies have proved that word representations from neural network language models such as BERT and GloVe have a gender tendency and convey gender information~\cite{jia2019gender,yang2020causal,lauscher2020general,Matthews_Hudzina_Sepehr_2022}. Among them,~\citet{jia2019gender} propose a BERT-based model in which each character, whether in or out of the sample, gets a representation from a pre-trained BERT that handles the `character-out-of-sample' problem for name-gender prediction. Besides, by concatenating the character embedding with the pronunciation embedding, their model, Pinyin BERT (PBERT), also proves that pronunciations deliver gender information. 

However, beyond semantics from characters themselves and their pronunciations, the semantics arisen from character components are overlooked in current gender guessing tools. A large portion of Chinese characters consists of components, and these components help shape the meanings of the characters~\cite{yu2017joint}. Following this lead, studies utilize the component-level internal semantic features for Chinese character representation learning in a word2vec fashion~\cite{sun2014radical,yu2017joint,zhang2019learning}. In this direction, \citet{wang2021improving} takes a step forward, capturing the semantic relationships between characters with shared components by constructing a homogeneous graph. In this way, characters and their components are inter-linked such that the semantic relationships between characters are better shaped and weighted with the attention mechanism. Though their FGAT model is the SOTA for many downstream NLP tasks, the relationships it relies on are in fact often heterogeneous, and it has not considered the same-pronunciation connections which can potentially augment the graph, as hinted by PBERT.

\begin{figure}[htb]
\includegraphics[width=5.5cm]{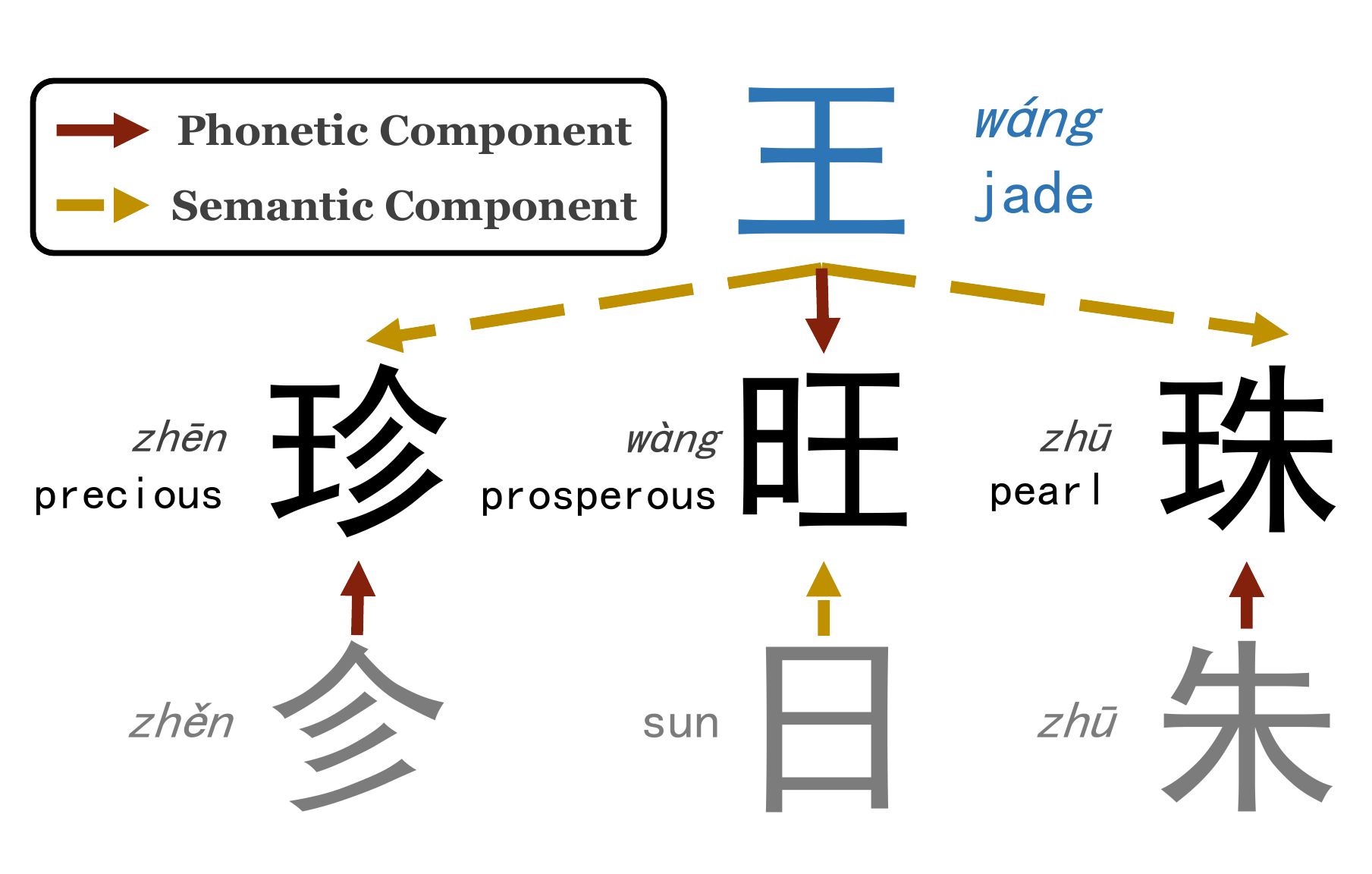}
\centering
\caption{Example of the shared-component connections.  `$\vcenter{\hbox{\includegraphics[width=0.27cm]{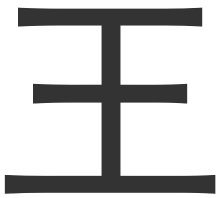}}}$' is the semantic component of `$\vcenter{\hbox{\includegraphics[width=0.29cm]{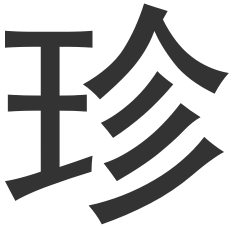}}}$' and `$\vcenter{\hbox{\includegraphics[width=0.29cm]{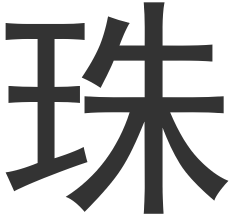}}}$', but is the phonetic component of `$\vcenter{\hbox{\includegraphics[width=0.26cm]{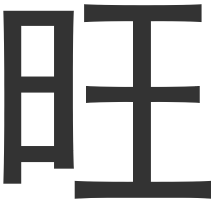}}}$'.}
\label{fig:shared-component}
\end{figure}

We illustrate the heterogeneity in shared-component connections with an example and discuss how it affects our model design. In Figure~\ref{fig:shared-component}, we see that the character `$\vcenter{\hbox{\includegraphics[width=0.29cm]{pictures/zhen1.png}}}$'  (precious, \textit{zh\={e}n})\footnote{The meaning and the pronunciation (italic) are in parentheses.}, `$\vcenter{\hbox{\includegraphics[width=0.29cm]{pictures/zhu1.png}}}$' (pearl, \textit{zh\={u}}) and `$\vcenter{\hbox{\includegraphics[width=0.26cm]{pictures/wang4.png}}}$' (prosperous, \textit{w\`{a}ng}) share the component `$\vcenter{\hbox{\includegraphics[width=0.27cm]{pictures/wang2.png}}}$' (jade, \textit{w\'{a}ng}). If modeled by a homogeneous graph as in FGAT, it will indicate that `$\vcenter{\hbox{\includegraphics[width=0.29cm]{pictures/zhen1.png}}}$', `$\vcenter{\hbox{\includegraphics[width=0.29cm]{pictures/zhu1.png}}}$', and `$\vcenter{\hbox{\includegraphics[width=0.26cm]{pictures/wang4.png}}}$' have equal pair-wised semantic similarity, contributed by the shared component `$\vcenter{\hbox{\includegraphics[width=0.27cm]{pictures/wang2.png}}}$'. Actually, `$\vcenter{\hbox{\includegraphics[width=0.29cm]{pictures/zhen1.png}}}$'  and `$\vcenter{\hbox{\includegraphics[width=0.29cm]{pictures/zhu1.png}}}$'  are much closer semantically, than to `$\vcenter{\hbox{\includegraphics[width=0.26cm]{pictures/wang4.png}}}$'. Regarding genders, `$\vcenter{\hbox{\includegraphics[width=0.29cm]{pictures/zhen1.png}}}$'  and `$\vcenter{\hbox{\includegraphics[width=0.29cm]{pictures/zhu1.png}}}$'  are popular in female names, while `$\vcenter{\hbox{\includegraphics[width=0.26cm]{pictures/wang4.png}}}$'  has a strong male tendency. Therefore, modeling their relationships homogeneously would mislead name-gender prediction. In fact, their relationships can be distinguished if we know the `$\vcenter{\hbox{\includegraphics[width=0.27cm]{pictures/wang2.png}}}$' in `$\vcenter{\hbox{\includegraphics[width=0.26cm]{pictures/wang4.png}}}$'  is a phonetic component (solid arrow in Figure~\ref{fig:shared-component}) which indicates the pronunciation of `$\vcenter{\hbox{\includegraphics[width=0.26cm]{pictures/wang4.png}}}$' and the `$\vcenter{\hbox{\includegraphics[width=0.27cm]{pictures/wang2.png}}}$' in `$\vcenter{\hbox{\includegraphics[width=0.29cm]{pictures/zhen1.png}}}$'  and `$\vcenter{\hbox{\includegraphics[width=0.29cm]{pictures/zhu1.png}}}$'  is a semantic component (dashed arrow in Figure~\ref{fig:shared-component}) contributing to its meaning. Besides `$\vcenter{\hbox{\includegraphics[width=0.27cm]{pictures/wang2.png}}}$', `$\vcenter{\hbox{\includegraphics[width=0.27cm]{pictures/wang4.png}}}$' also has a semantic component `$\vcenter{\hbox{\includegraphics[width=0.27cm]{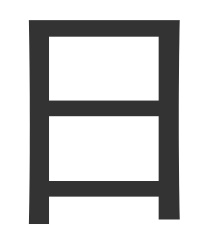}}}$' (sun) as shown at the bottom of Figure~\ref{fig:shared-component}. Similarly, `$\vcenter{\hbox{\includegraphics[width=0.29cm]{pictures/zhen1.png}}}$' and `$\vcenter{\hbox{\includegraphics[width=0.29cm]{pictures/zhu1.png}}}$' has `$\vcenter{\hbox{\includegraphics[width=0.27cm]{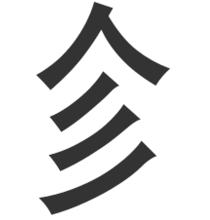}}}$' (\textit{zh\v{e}n}) and '$\vcenter{\hbox{\includegraphics[width=0.30cm]{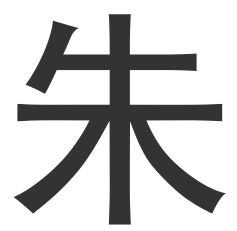}}}$' 
(\textit{zh\={u}})
as their phonetic components, respectively. A character with both semantic component and phonetic component, such as `$\vcenter{\hbox{\includegraphics[width=0.29cm]{pictures/zhen1.png}}}$', `$\vcenter{\hbox{\includegraphics[width=0.29cm]{pictures/zhu1.png}}}$' and `$\vcenter{\hbox{\includegraphics[width=0.26cm]{pictures/wang4.png}}}$', is called a picto-phonetic character. 80.5\% of Chinese characters are picto-phonetic~\cite{sun1997}, suggesting the effect of the heterogeneity in shared-component relationships is non-negligible. Therefore, we design heterogeneous graphs that specify character-semantic component edges and character-phonetic component edges.

Within this model structure, we introduce the shared-pronunciation connection as a new type of edges, given the effectiveness of pronunciations in gender guessing. As a straightforward example, characters sharing the pronunciation `mei' (the same pronunciation of the character `$\vcenter{\hbox{\includegraphics[width=0.27cm]{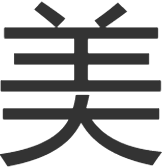}}}$', which means beautiful) are more likely to be in female names. Now with character-semantic component edges, character-phonetic component edges, and character-pronunciation edges, we use a multi-level Chinese character attention network first to learn the importance of different components and structural information at the component-level and then aggregates the gender information conveyed by pronunciations (i.e., pinyin).

In addition to methodological contributions, we provide a high-quality Chinese name-gender dataset. Most Chinese name-gender datasets, such as the Ngender dataset, only contains frequencies of individual characters instead of complete names and loses important information for Chinese name-gender prediction.
Unlike English first names, which are usually one-word names, most Chinese first names have one or two characters, with two-character names being the majority (84.55\%)\footnote{www.mps.gov.cn/n2253534/n2253535/c8349222/content.html}. For Chinese first names with more than one character, the combinations of characters can be informative, and they sometimes even deliver opposite information as we can get from individual characters. For instance, `$\vcenter{\hbox{\includegraphics[width=0.27cm]{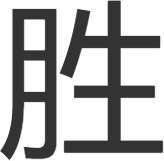}}}$'(win) and `$\vcenter{\hbox{\includegraphics[width=0.25cm]{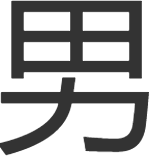}}}$’(male) are characters appearing more in male names, but `$\vcenter{\hbox{\includegraphics[width=0.27cm]{pictures/sheng.png}}}$ $\vcenter{\hbox{\includegraphics[width=0.25cm]{pictures/nan.png}}}$'(triumph over males) is a female name. Furthermore, character combinations $A$-$B$ and $B$-$A$ sometimes differ in gender probabilities, as suggested by our analysis in the Dataset section. 
To unleash the power of sequential representations and promote relevant research, we open source our large-scale full first name data, collected from an official source.

To sum up, our contributions in this paper include:
\begin{enumerate}
    \item To the best of our knowledge, this is the first work that uses a graph neural network to model Chinese characters' internal and external connections for name-gender prediction to facilitate gender-equality studies.
    \item We propose a heterogeneous graph with a multi-level attention network to capture the heterogeneity in semantic relationships between characters and components, as well as gender inclination indicated by pronunciations. The model outperforms various baselines and reaches a state-of-the-art accuracy of 93.62\%.
    \item We provide a dataset of 58 million Chinese names with associated genders as well as our source code\footnote{\url{https://github.com/ZhangDataLab/CHGAT}}, in hopes of promoting gender equality research, especially for the underrepresented Chinese females.
\end{enumerate}

\section{Related Work}
\subsection{Name-gender Prediction}
Name-gender prediction is a common task in gender-quality studies~\cite{lariviere2013bibliometrics,huang2020historical,koning2021we}, as well as in web-based services like advertising and recommendation systems~\cite{mukherjee2017gender,wu2019neural}. Names, being very informative in many languages and cultures, are one important clue for gender guessing.

For western names, \citet{wais2016gender} designs genderizeR to predict the gender of an input name, simply according to the majority gender of people under this name in their data. Neural network language models further help exploit more information and tackle the `out-of-sample' problem. \citet{hu2021s} construct a character-level BERT-based model to guess genders from English names.

Chinese, different from Latin-based languages, is logo graphic.  Besides the Chinese characters, their components and pronunciations also convey information~\cite{cao2018cw2vec}. Furthermore, in multi-character Chinese names, combinations across characters bring additional information. However, most gender-guessing tools only rely on character-level information. For instance, Ngender and the model proposed by \citet{zhao2017advance} both regard the product of all characters' probabilities under a gender as the name's probability of being that gender, and output the gender with the highest probability. They have natural limitations in capturing the extra information that comes with the combinations of characters. \citet{jia2019gender} concatenate character embeddings and pronunciation embeddings from pre-trained BERT models and mitigate these limitations. However, the connections between Chinese characters indicated by shared components and pronunciations are yet to be fully exploited.

\subsection{Representing Characters with Components}
 The components of Chinese characters convey rich semantic information.
Previous work incorporates the component information into the character embeddings based on the word2vec model to learn the representation of the Chinese characters~\cite{sun2014radical,yin2016multi,yu2017joint}. Seeking finer-grained information, some studies break components into sequenced strokes (subcomponents) to enhance the representations~\cite{cao2018cw2vec,zhang2019learning}. Though the characters are decomposed into components or subcomponents, the word2vec model cannot discriminate their importance and irrelevant parts can introduce noises into the representations.

Hence, \citet{wang2021improving} model a character and its components in a homogenous graph with attention to learn the importance of components. Meanwhile, it shapes the character semantics through the characters' connections with other characters sharing the same components. As a result, their FGAT model achieves SOTA results on various downstream NLP tasks. However, as discussed in Introduction, their homogenous graph cannot model the heterogeneity in character-component relationships, since the majority of Chinese characters are picto-phonetic and the character-semantic component relationships and character-phonetic component relationships should be specified individually in the form of heterogeneous graphs. As an add-on, the shared-pronunciation relationships can also be integrated in heterogeneous graphs, potentially bringing more gender information for our prediction task.
   
\section{Method}
The overall structure of our model is shown in Figure~\ref{fig:structure}.
We take the Chinese characters and their pronunciations as the inputs to learn the information from the intra-character and intra-pronunciation combinations in names. Input names in the form of Chinese characters go into a Chinese Heterogeneous Graph Attention (CHGAT) layer and the BERT text encoder layer simultaneously. The output embeddings from the two layers are added and concatenated with the name's pronunciation embedding, generated by the BERT text encoder layer. This embedding is then fed into the Transformer encoder module to learn the contextual information within names. Finally, the classifier, a single layer fully connected network, is used to predict the gender. We detail the design of our heterogeneous graph as well as its core component, the CHGAT layer, in the following sections.

\subsection{Heterogeneous Graph Structure}
\subsubsection{Formation of Chinese Characters.}
Chinese evolves from an ancient hieroglyph writing system. Basic characters such as wood and fire appeared first, with their individual graphical representations. Characters indicating more complex concepts, are composed by the combinations of these basic characters~\cite{tao2019radical}.

For example, by paralleling two `$\vcenter{\hbox{\includegraphics[width=0.29cm]{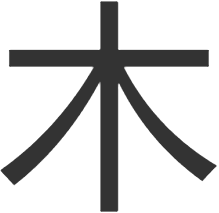}}}$ (wood, \textit{m\={u}})' we get `$\vcenter{\hbox{\includegraphics[width=0.29cm]{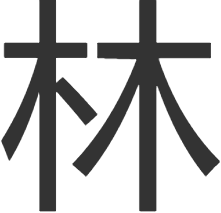}}}$  (woods, \textit{lín})'. When `$\vcenter{\hbox{\includegraphics[width=0.29cm]{pictures/lin.png}}}$  (woods, \textit{lín})' is further combined with `$\vcenter{\hbox{\includegraphics[width=0.29cm]{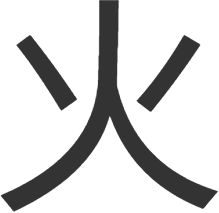}}}$ (fire, \textit{hu\v{o}})' underneath it, it becomes `$\vcenter{\hbox{\includegraphics[width=0.29cm]{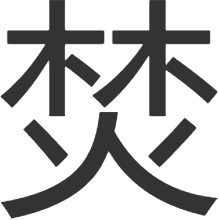}}}$ (burn, \textit{f\'{e}n})'.

Although the shapes of characters change over time, how they are combined (i.e., their structures) are preserved. There are 17 types of structures in modern Chinese according to QXK\footnote{https://qxk.bnu.edu.cn/\#/help} (a Chinese character information system built by researchers from Beijing Normal University) and we list them in Table~\ref{tab:structure}. As mentioned in Introduction, 80.5\% of simplified Chinese characters are picto-phonetic, and QXK provides the semantic components and phonetic components of most Chinese characters.

\begin{table}[htb]
\small
\centering
\begin{spacing}{1.2}
\begin{tabular}{|c|c|c|c|}
\hline
Index & Structure Type            & Icon & Example \\ \hline
1     & left to right             & $\vcenter{\hbox{\includegraphics[width=0.29cm]{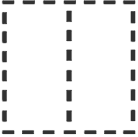}}}$    & $\vcenter{\hbox{\includegraphics[width=0.29cm]{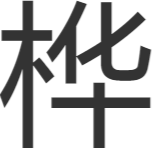}}}$, $\vcenter{\hbox{\includegraphics[width=0.29cm]{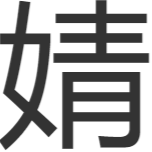}}}$        \\  \hline
2     & left to middle and right  & $\vcenter{\hbox{\includegraphics[width=0.29cm]{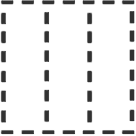}}}$    &    $\vcenter{\hbox{\includegraphics[width=0.29cm]{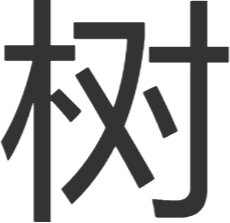}}}$, $\vcenter{\hbox{\includegraphics[width=0.29cm]{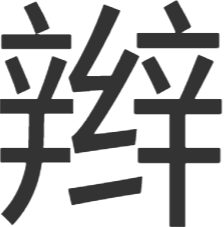}}}$      \\  \hline
3     & above to below            & $\vcenter{\hbox{\includegraphics[width=0.29cm]{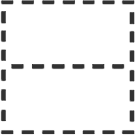}}}$    &    $\vcenter{\hbox{\includegraphics[width=0.29cm]{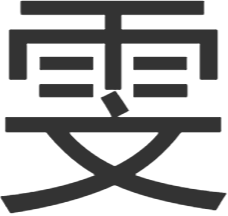}}}$, $\vcenter{\hbox{\includegraphics[width=0.29cm]{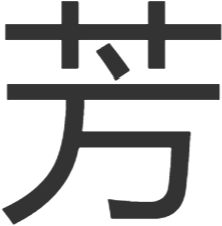}}}$      \\  \hline
4     & above to middle and below & $\vcenter{\hbox{\includegraphics[width=0.29cm]{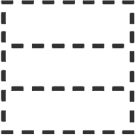}}}$    &     $\vcenter{\hbox{\includegraphics[width=0.29cm]{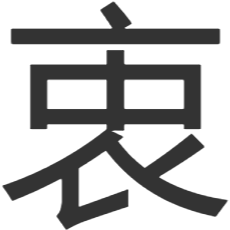}}}$, $\vcenter{\hbox{\includegraphics[width=0.29cm]{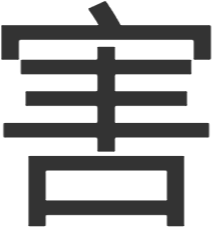}}}$     \\  \hline
5     & full surround             & $\vcenter{\hbox{\includegraphics[width=0.29cm]{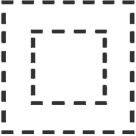}}}$    &     $\vcenter{\hbox{\includegraphics[width=0.29cm]{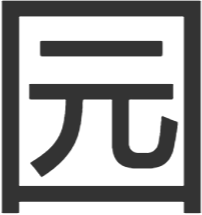}}}$, $\vcenter{\hbox{\includegraphics[width=0.29cm]{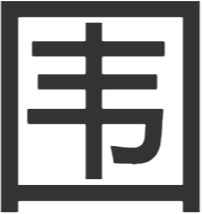}}}$     \\  \hline
6     & surround from above       & $\vcenter{\hbox{\includegraphics[width=0.29cm]{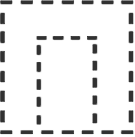}}}$    &     $\vcenter{\hbox{\includegraphics[width=0.29cm]{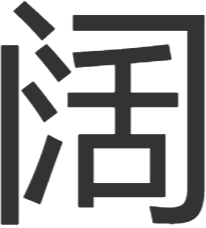}}}$, $\vcenter{\hbox{\includegraphics[width=0.29cm]{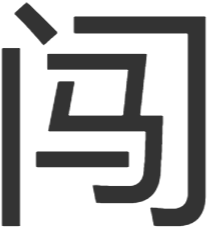}}}$     \\  \hline
7     & surround from below       & $\vcenter{\hbox{\includegraphics[width=0.29cm]{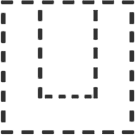}}}$    &     $\vcenter{\hbox{\includegraphics[width=0.29cm]{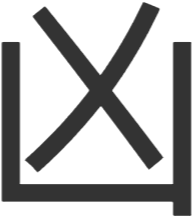}}}$, $\vcenter{\hbox{\includegraphics[width=0.29cm]{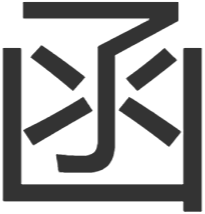}}}$     \\  \hline
8     & surround from left        & $\vcenter{\hbox{\includegraphics[width=0.29cm]{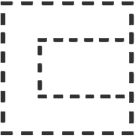}}}$    &    $\vcenter{\hbox{\includegraphics[width=0.29cm]{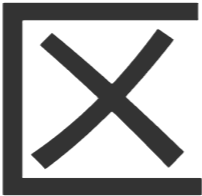}}}$, $\vcenter{\hbox{\includegraphics[width=0.29cm]{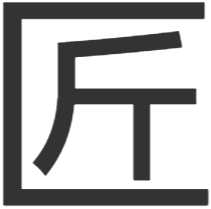}}}$      \\  \hline
9     & surround from upper left  & $\vcenter{\hbox{\includegraphics[width=0.29cm]{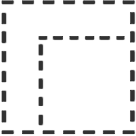}}}$    &    $\vcenter{\hbox{\includegraphics[width=0.29cm]{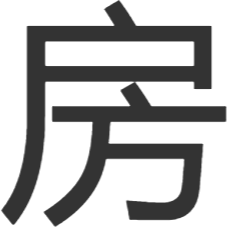}}}$, $\vcenter{\hbox{\includegraphics[width=0.29cm]{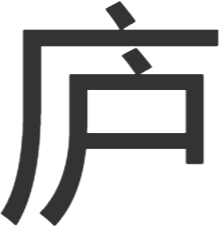}}}$      \\  \hline
10    & surround from upper right & $\vcenter{\hbox{\includegraphics[width=0.29cm]{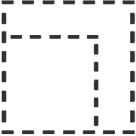}}}$    &    $\vcenter{\hbox{\includegraphics[width=0.29cm]{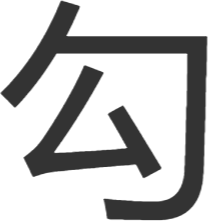}}}$, $\vcenter{\hbox{\includegraphics[width=0.29cm]{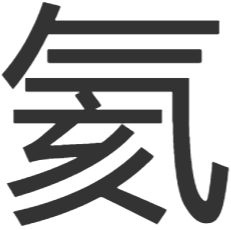}}}$      \\  \hline
11    & surround from lower left  & $\vcenter{\hbox{\includegraphics[width=0.29cm]{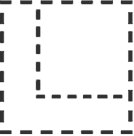}}}$    &    $\vcenter{\hbox{\includegraphics[width=0.29cm]{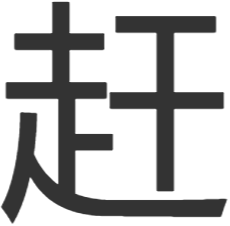}}}$, $\vcenter{\hbox{\includegraphics[width=0.29cm]{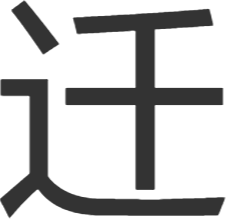}}}$      \\  \hline
12    & integral                  & $\vcenter{\hbox{\includegraphics[width=0.29cm]{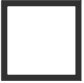}}}$    &    $\vcenter{\hbox{\includegraphics[width=0.29cm]{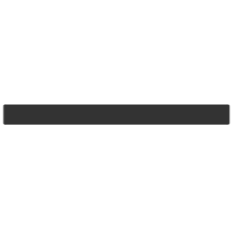}}}$, $\vcenter{\hbox{\includegraphics[width=0.29cm]{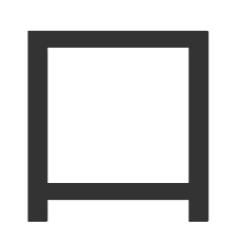}}}$      \\  \hline
13    & isosceles triangle layout & $\vcenter{\hbox{\includegraphics[width=0.29cm]{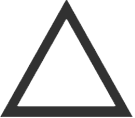}}}$    &    $\vcenter{\hbox{\includegraphics[width=0.29cm]{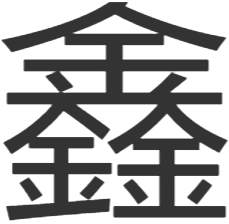}}}$, $\vcenter{\hbox{\includegraphics[width=0.29cm]{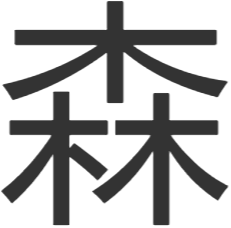}}}$      \\  \hline
14    & square layout             & $\vcenter{\hbox{\includegraphics[width=0.29cm]{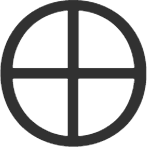}}}$    &    $\vcenter{\hbox{\includegraphics[width=0.29cm]{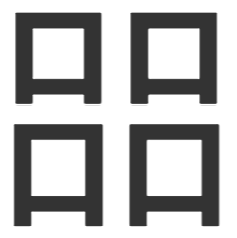}}}$, $\vcenter{\hbox{\includegraphics[width=0.29cm]{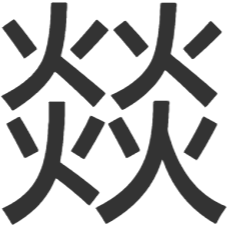}}}$      \\  \hline
15    & multielement combination  & $\vcenter{\hbox{\includegraphics[width=0.29cm]{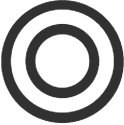}}}$    &    $\vcenter{\hbox{\includegraphics[width=0.29cm]{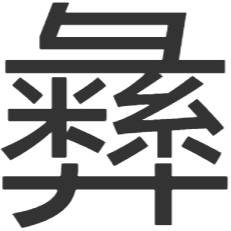}}}$, $\vcenter{\hbox{\includegraphics[width=0.29cm]{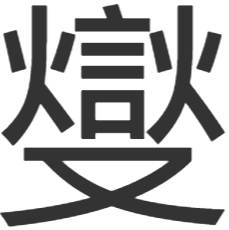}}}$      \\  \hline
16    & overlaid                  & $\vcenter{\hbox{\includegraphics[width=0.29cm]{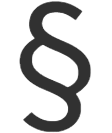}}}$    &    $\vcenter{\hbox{\includegraphics[width=0.29cm]{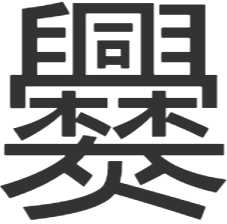}}}$, $\vcenter{\hbox{\includegraphics[width=0.29cm]{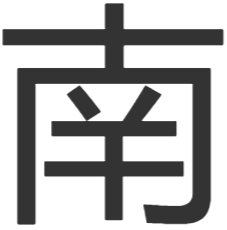}}}$      \\  \hline
17    & multielement stacking     & $\vcenter{\hbox{\includegraphics[width=0.29cm]{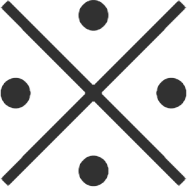}}}$    &    $\vcenter{\hbox{\includegraphics[width=0.29cm]{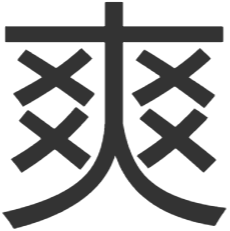}}}$, $\vcenter{\hbox{\includegraphics[width=0.29cm]{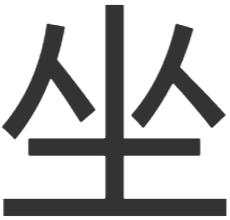}}}$      \\  \hline
\end{tabular}
\end{spacing}
\caption{The list of Chinese characters' formation types. The first 11 structure type names follow their unicode names, while the last 6 have no formal names, so we describe their structural characteristics as best we can.}
\label{tab:structure}
\end{table}

\begin{figure*}[htb]
\includegraphics[width=0.85\linewidth]{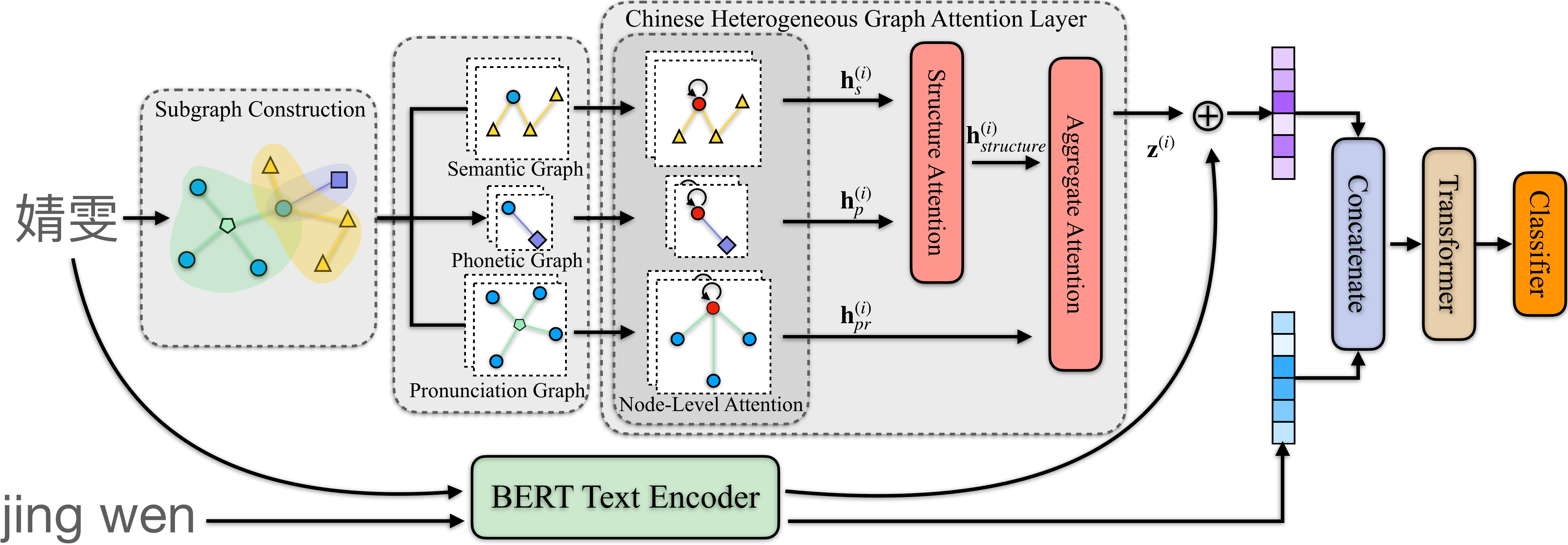}
\centering
\caption{System structure.}
\label{fig:structure}
\end{figure*}

\subsubsection{Graph Structure.}
In this paper, we define a heterogeneous graph with four types of nodes, namely, \textbf{\textit{character}}, \textbf{\textit{semantic component}},  \textbf{\textit{phonetic component}}, and \textbf{\textit{pronunciation}}, and three types of paths, including \textbf{\textit{character-semantic component}}, \textbf{\textit{character-phonetic component}}, and \textbf{\textit{character-pronunciation}}, to incorporate gender information.

If a character is picto-phonetic, it connects with its one component representing the sound through a character-phonetic component edge. For its other components, it connects with them by character-semantic component edges. If a character is non-picto-phonetic, it connects with its components through character-semantic component edges as well. Besides, a character-pronunciation edge is added between the character and its pronunciation.

\begin{figure}[!hptb]
\includegraphics[width=0.9\linewidth]{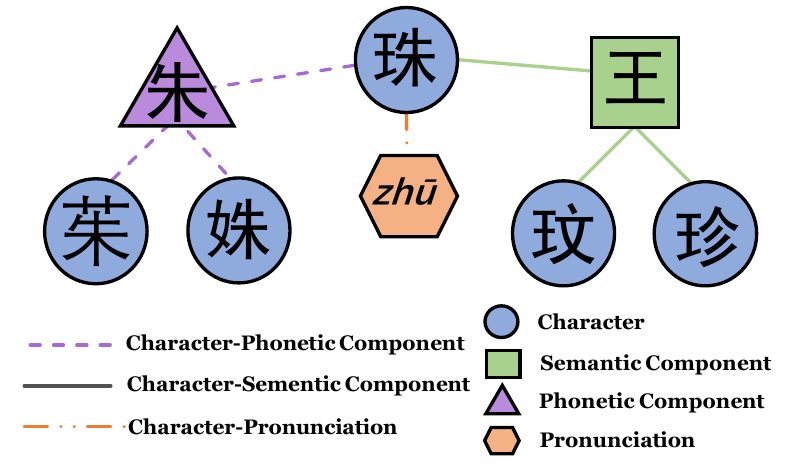}
\centering
\captionof{figure}{An example of graph composition of the character `$\vcenter{\hbox{\includegraphics[width=0.26cm]{pictures/zhu1.png}}}$'.
`$\vcenter{\hbox{\includegraphics[width=0.27cm]{pictures/zhu_pian.png}}}$', `zh\={u}' and `$\vcenter{\hbox{\includegraphics[width=0.26cm]{pictures/wang2.png}}}$' are the phonetic component, pronunciation and semantic compoente of `$\vcenter{\hbox{\includegraphics[width=0.26cm]{pictures/zhu1.png}}}$', and they are connected with `$\vcenter{\hbox{\includegraphics[width=0.26cm]{pictures/zhu1.png}}}$' through the character-phonetic component, character pronunciation and character-sementic component, respectively.}
\label{fig:example graph}
\end{figure}

We take an example of the character 
`$\vcenter{\hbox{\includegraphics[width=0.26cm]{pictures/zhu1.png}}}$' to demonstrate how characters, pronunciations, and different types of components are represented and connected in our heterogeneous graph.
In Figure~\ref{fig:example graph}, the focal character 
`$\vcenter{\hbox{\includegraphics[width=0.26cm]{pictures/zhu1.png}}}$'
and its phonetic component 
`$\vcenter{\hbox{\includegraphics[width=0.26cm]{pictures/zhu_pian.png}}}$'
are connected  by a character-phonetic component edge, and character 
`$\vcenter{\hbox{\includegraphics[width=0.26cm]{pictures/zhu1.png}}}$'
and its semantic component 
`$\vcenter{\hbox{\includegraphics[width=0.26cm]{pictures/wang2.png}}}$'
are connected by a character-semantic component edge.

 Previous research~\cite{wang2021improving} utilizes the homogeneous graph to formulate the relations of components to characters, where there is only one type of node representing all components and characters and one type of edge for inter-character shared-component relationships. They assume that a glyph represents the same semantics when used as a character, a semantic component, or a phonetic component, which is oversimplified. 
 We distinguish between the semantic and phonetic components, and connect the focal character to all the characters sharing the same components.

As discussed in Introduction section, \citet{jia2019gender} show that character pronunciations have gender inclinations, but they only use pronunciations as independent representations, without the message interacting with character embeddings. Hence, we connect the character  
`$\vcenter{\hbox{\includegraphics[width=0.26cm]{pictures/zhu1.png}}}$'
and its pronunciation `\textit{zh\={u}}' by a character-pronunciation edge, as shown in Figure~\ref{fig:example graph}.

\subsection{Chinese Heterogeneous Graph Attention}
To aggregate the information in the heterogeneous graph, we design the Chinese Heterogeneous Graph Attention (CHGAT) layer with three-level attentions. Both the semantic graph and the phonetic graph contain structural information. After the node-level attention aggregates the information within each graph,  structural information remains scattered among \textit{semantic component} and \textit{phonetic component} nodes. Therefore, the network captures the structural information of a character by aggregating the two node-level representations. To assemble information from the pronunciation and the structure representations, another attention, i.e., the aggregate attention, is used.

We denote the set of Chinese characters as $\mathcal{C}=\{c_0,c_1,...,c_m\}$. 
A character $c_i$'s feature embedding is $\boldsymbol{f}^{(i)}_c$.
All characters share one pronunciation graph $\boldsymbol{g}_{pr}$, which is a `character-pronunciation-character' meta-path in~\cite{wang2019heterogeneous}'s definition, but each character has its own semantic graph and phonetic graph. For a character $c_i$, its semantic graph contains all the one-hop and two-hop semantic components of $c_i$, denoted as $\boldsymbol{g}^{(i)}_s$, while its phonetic graph is the one containing the one-hop phonetic component of $c_i$, represented by $\boldsymbol{g}^{(i)}_p$. The feature embedding of component $s_o$ in the semantic graph is $\boldsymbol{f}^{(o)}_s$, while the feature embedding of component $p_t$ in the phonetic graph is $\boldsymbol{f}^{(t)}_p$.

\citet{wang2021improving} introduce position embedding that adds position information to the component representations. The same components may appear in various positions of different characters, and this position information can affect the meanings of these components. Hence, we add another position embedding, which represents the position of each component in the character, denoted as
\begin{align}
    \boldsymbol{\lambda}=\{\boldsymbol{\lambda}^{(0)},\boldsymbol{\lambda}^{(1)},...,\boldsymbol{\lambda}^{(a)}\}.
\end{align}

Therefore, the initial embedding $\boldsymbol{x}_c^i$, the semantic components' initial embedding $\boldsymbol{x}_s^o$, and the phonetic components' initial embeddings $\boldsymbol{x}_p^t$ for an input character are:

\begin{equation}
    \boldsymbol{x}_{c}^{i} = \boldsymbol{f}_{c}^{(i)} + \boldsymbol{\lambda}^{(l_c)},
\end{equation}

\begin{equation}
    \boldsymbol{x}_{s}^{o} = \boldsymbol{f}_{s}^{(o)} + \boldsymbol{\lambda}^{(l_s)}.
\end{equation}

\begin{equation}
    \boldsymbol{x}_{p}^{t} = \boldsymbol{f}_{p}^{(t)} + \boldsymbol{\lambda}^{(l_p)},
\end{equation}
respectively. $i$, $o$, and $t$ denote the index of character, semantic component, and phonetic component, respectively. $l_c$, $l_s$, and $l_p$ represent their corresponding position index, respectively.

Here, we define character $c_i$'s phonetic graph feature embeddings as $\boldsymbol{X}^{(i)}_p$, and its semantic graph feature embeddings as $\boldsymbol{X}^{(i)}_s$. The pronunciation graph feature embedding is $\boldsymbol{X}^{(i)}_{pr}$.

Inspired by HAN~\cite{wang2019heterogeneous}, we use a three-level attention mechanism in our scenario. As illustrated in Figure~\ref{fig:structure}. The model first learns the node-level embeddings within each graph, and then the node-level embeddings of a semantic graph and phonetic graph are fed into the structure attention to learn a structure embedding.
This embedding and the node-level pronunciation embedding are used to learn the character's final embedding with an aggregated attention.

\subsubsection{Node-level Attention.}
We first project different types of node features into the same feature space with a transformation matrix:
\begin{equation}
    \boldsymbol{\gamma}_{k}^{(i)} = \boldsymbol{W}_{k}^{(i)} \boldsymbol{x}_{k}^{(i)},
\end{equation}
where $\boldsymbol{W}_{k}^{(i)} \in \mathbb{R}^{d_k \times d}$ is a learnable parameter. $k \in \{s,p,pr\}$ represents path type, and $d_k$ is the input feature dimension. The importance score of node $j$ to target node $i$ is computed as:

\begin{equation}
n_k^{(i,j)} =  \boldsymbol{LeakyReLU}(\boldsymbol{w}_k[\boldsymbol{\gamma}_{k}^{(i)}||\boldsymbol{\gamma}_{k}^{(j)}]),
\end{equation}
where $\boldsymbol{w}_k \in \mathbb{R}^{1\times 2d}$ is a learnable vector. The importance score is then normalized to be the weight of node $j$ to node $i$, denoted as $\theta_k^{(i,j)}$:

\begin{equation}
\theta_k^{(i,j)} = \frac{\exp(n_k^{(i,j)})}{\sum_{j \in \mathcal{N}_k^{(i)}} \exp(n_k^{(i,j)})},
\end{equation}
where $\mathcal{N}_k^{(i)}$ is the set of all $i$'s neighbors in the $k$ type of path. The node-level embedding is then the weighted sum of all nodes connecting to itself:

\begin{equation}
\boldsymbol{h}_k^{(i)} = \underset{t}{\mathrm{\|}}\boldsymbol{ELU}(\sum_{j\in \mathcal{N}_k^{(i)}} \theta^{(i,j)}\boldsymbol{\gamma}_k^{(i)}),
\end{equation}
where $t$ is the number of heads, and $\mathrm{\|}$ represents the concatenation operation.

\subsubsection{Attention Module.}
The attention module aggregates the representation with different semantics into one representation. We denote it as:
\begin{equation}
\boldsymbol{h}^{(i)} = attn(\boldsymbol{h}_{1}^{(i)},\boldsymbol{h}_{2}^{(i)},...,\boldsymbol{h}_{v}^{(i)}),
\end{equation}
where $v$ represents the number of inputs. The importance of each input to the target embedding is:
\begin{equation}
w_r =  \frac{1}{|N|} \sum_{i\in N}\boldsymbol{q}^T( tanh(\boldsymbol{W}\boldsymbol{h}_{r}^{(i)}+\boldsymbol{b})),
\end{equation}
where $\boldsymbol{q}$, $\boldsymbol{W}$ and $\boldsymbol{b}$ are learnable parameters in the model. $r \in \{1,2,...,v\}$ is the type of semantics, and $N$ is the input name. Then the importance score of each input is:
\begin{equation}
\delta{r} = \frac{exp(w_{r})}{\sum_{u\in[1,v]}exp(w_{u}) }.
\end{equation}
The target embedding is:
\begin{equation}
\boldsymbol{h}^{(i)} = \sum_{r\in [1,v]}\delta{r} \boldsymbol{h}_{r}^{(i)}.
\end{equation}

\subsubsection{Structure Attention Layer.}
In the structure attention layer, we learn the structure representation of $c_{i}$ from its node-level embeddings of the semantic graph and the phonetic graph with an attention module:
\begin{equation}
\boldsymbol{h}_{structure}^{(i)} = attn(\boldsymbol{h}_{s}^{(i)}, \boldsymbol{h}_{p}^{(i)}).
\end{equation}
The $\boldsymbol{h}_{structure}^{(i)}$ denotes the structure representation of character $c_{i}$.

\subsubsection{Aggregate Attention Layer.}
The aggregate attention layer assembles the character $c_{i}$' pronunciation representation, and its structure representation into one embedding. Again, this is achieved by applying an attention module, which is:
\begin{equation}
\boldsymbol{z}^{(i)} = attn(\boldsymbol{h}_{structure}^{(i)}, \boldsymbol{h}_{pr}^{(i)}).
\end{equation}
Finally, $\boldsymbol{z}^{(i)}$ is the output of $c_i$ at the CHGAT layer.
\subsection{Loss Function}
Our objective function is defined as:
\begin{equation}
L = -\sum^J_j y_jlog(\hat{y}_{j})+(1-y_j)log(1-\hat{y}_{j})
\end{equation}
where $J$ represents the number of data, $y_j$ is the label of the $j^{th}$ input, and $\hat{y}_{j}$ represents the predicted probability of $y_j$ being the label.

\section{Dataset}
\label{sec:dataset}
\begin{table}[!hbt]
\small
  \centering
  \scalebox{0.95}{
    \begin{tabular}{lccc}
    \toprule
    &\multicolumn{1}{c}{Records}&\multicolumn{1}{c}{Unique First Names}&\multicolumn{1}{c}{M-to-F\%}\\
\hline

Ngender&     32,067,566&      9,442&  197.28\\

Our Dataset&     58,393,173&      560,706&  111.58  \\

9,800 Names&     9,800&      6,972& 100.00 \\

25,856 Names&     25,856&      21,051& 100.00 \\

    \bottomrule
    \end{tabular}%
    }
    \caption{Statistics of the datasets. For the Ngender dataset, we count its unique characters as its unique first names, since it is a single-character dataset. Besides, we show the ratio of males to females for each dataset.}
    \label{tab: dataset describtion}
\end{table}%

We provide a dataset with 58,393,173 records of 560,706 different first names and the associated gender for each name occurrence,  collected from an official source.
To be specific, we begin with 8,224,820 unique full names without gender information in a company registration dataset from China's State Administration for Industry and Commerce. These full names are then used as queries to a service from Guangdong province government that provides the number of females and males with the querying name. The resulting gender frequencies are aggregated under the 560,706 unique first names to form our dataset.As a comparison, the most popular Chinese name-gender prediction tool (Ngender) on GitHub provides a dataset of 32,067,566 entries of 9,442 characters which is collected from unofficial sources. 
The detail information of the datasets is shown in Table \ref{tab: dataset describtion}. 

Our dataset is naturally more informative than the Ngender dataset -- Ngender only provides for each character the numbers of females and males with this particular character in their first names, and the information from multi-character combinations is absent. However, this combination conveys important gender information. For instance, when two characters associated with the same gender are put together, their combination can indicate the opposite gender, as discussed in Introduction. According to our statistics, these cases account for 1.75\% of names in our dataset.
Beyond this, we discover that when we reverse the two characters in names, 14.77\% of names would have reversed gender tendency. These confirm that character combinations actually deliver helpful information for distinguishing genders.

Moreover, our dataset is more balanced with a male-to-female ratio (M-to-F) of 111.59\%, against a highly unbalanced ratio of 197.28\% for Ngender.

\section{Experiments}
\subsection{Experimental Setup}
\subsubsection{Datasets.}
Besides our dataset, we use the aforementioned Ngender dataset to train the models as a comparison for enhancements from improved data quality. We test the models not only by splitting training and test sets, but also by introducing test data from two independent sources containing names and genders. We name them \textbf{9,800 Names}~\cite{9800dataset} and \textbf{25,856 Names}~\cite{du2020quantified} respectively. Their detailed information is in Table~\ref{tab: dataset describtion}.

\subsubsection{Implementation Details.}
Both the Ngender dataset and our dataset are split into 90\% training, 5\% validation, and 5\% test. All models are trained with the same epochs. The learning rate and the weight decay value of each model are adjusted with grid search. We use the accuracy score to evaluate all models, which is the number of correctly guessed instances over the total instances.

The initial embeddings in all models are randomly initialized. For all tasks, we set the number of attention heads to 6 and the dimension of embedding vectors to 768. We use AdamW as the optimizer. The learning rate and the weight decay of all models are adjusted with grid search.

\subsubsection{Baselines.}
We compare our method (CHGAT) with three representative baselines:
\begin{itemize}
    \item \textbf{Ngender:} A commonly used Chinese name-gender prediction tool based on Naïve Bayes.
    \item \textbf{Pinyin BERT~\cite{jia2019gender}:} Pinyin BERT (PBERT) makes use of characters' semantics from a pre-trained BERT as well as the gender information delivered in pronunciations by concatenating the two embeddings and feeding them into a BERT model.
    \item \textbf{FGAT~\cite{wang2021improving}:} The Chinese character formation graph attention network is a state-of-the-art model in Chinese character representation learning. It is a multitask representation model that uses a homogeneous graph to capture the semantic information delivered by a character's components. To make a fair comparison, we include pronunciation by concatenating the name's pronunciation embedding to the output character embedding from FGAT, and use the concatenated embedding to predict name gender.
\end{itemize}

\begin{table*}[htb]
\small
  \centering

    \begin{tabular}{cccccc}

    \toprule
\multirow{3}{*}{ \makecell{}}&\multirow{3}{*}{} &
\multicolumn{4}{c}{Testing Dataset} \\\cmidrule(lr){3-6}

  \makecell[c]{Training} & \makecell[c]{Method}  & \makecell[c]{9,800 Names}   & \makecell[c]{25,856 Names} & Ngender  & Ours  \\
    \midrule
\multirow{4}{*}{\makecell{Ours}}&Ngender&0.7066&0.7529&0.6636&0.8757\\
    &PBERT&0.8136&0.8081&0.7849&0.9309\\
    &FGAT&0.8139&0.8126&0.7854&0.9329\\
    &CHGAT&\textbf{0.8147}&\textbf{0.8186}&\textbf{0.7873}&\textbf{0.9362}\\
    \midrule
\multirow{4}{*}{\makecell{Ngender\\ }}&Ngender&0.6868&0.7518&0.6636&0.8476\\
    &PBERT&0.7541&0.7776&0.8010&0.9012\\
    &FGAT&0.7798&0.7977&0.8042&0.9148\\
    &CHGAT&\textbf{0.7897}&\textbf{0.8040}&\textbf{0.8054}&\textbf{0.9168}\\
    \bottomrule
    \end{tabular}%
  
    \caption{Experiment results of all models trained on the Ngender data and our data, and tested on four datasets.}
    \label{tab: Method result}%
\end{table*}%

\subsection{Experiment Results}
%
Table~\ref{tab: Method result} shows that our model hits the highest accuracy scores in all training and test combinations. When the experiment is conducted entirely on our dataset, it achieves the accuracy of 93.62\% which is significantly higher than the public available Ngender on its public dataset (84.76\%). 

Our method and FGAT, as the graph-based methods, outperform PBERT in all experiments by up to 4.72\% relatively. This indicates the structural information captured plays an important part in name-gender guessing. Our method further surpasses the current SOTA, FGAT, by 0.14\% to 1.27\% when trained on the Ngender dataset,  suggesting the heterogeneity in component relationships and information conveyed in shared pronunciations help the prediction.

It is worth noting that models trained on our dataset all outperform themselves trained on the Ngender dataset, with exceptions when the test set is from Ngender (the `Ngender' column in Table~\ref{tab: Method result}). This suggests that our dataset is a better source for training data and a possible reason for the exceptions is that dataset-dependant information may be learned, and it helps predict the test examples from the same dataset.

\subsection{Ablation Study}
\label{sec:ablation study}

\begin{figure}[hbt]
\includegraphics[width=0.6\linewidth]{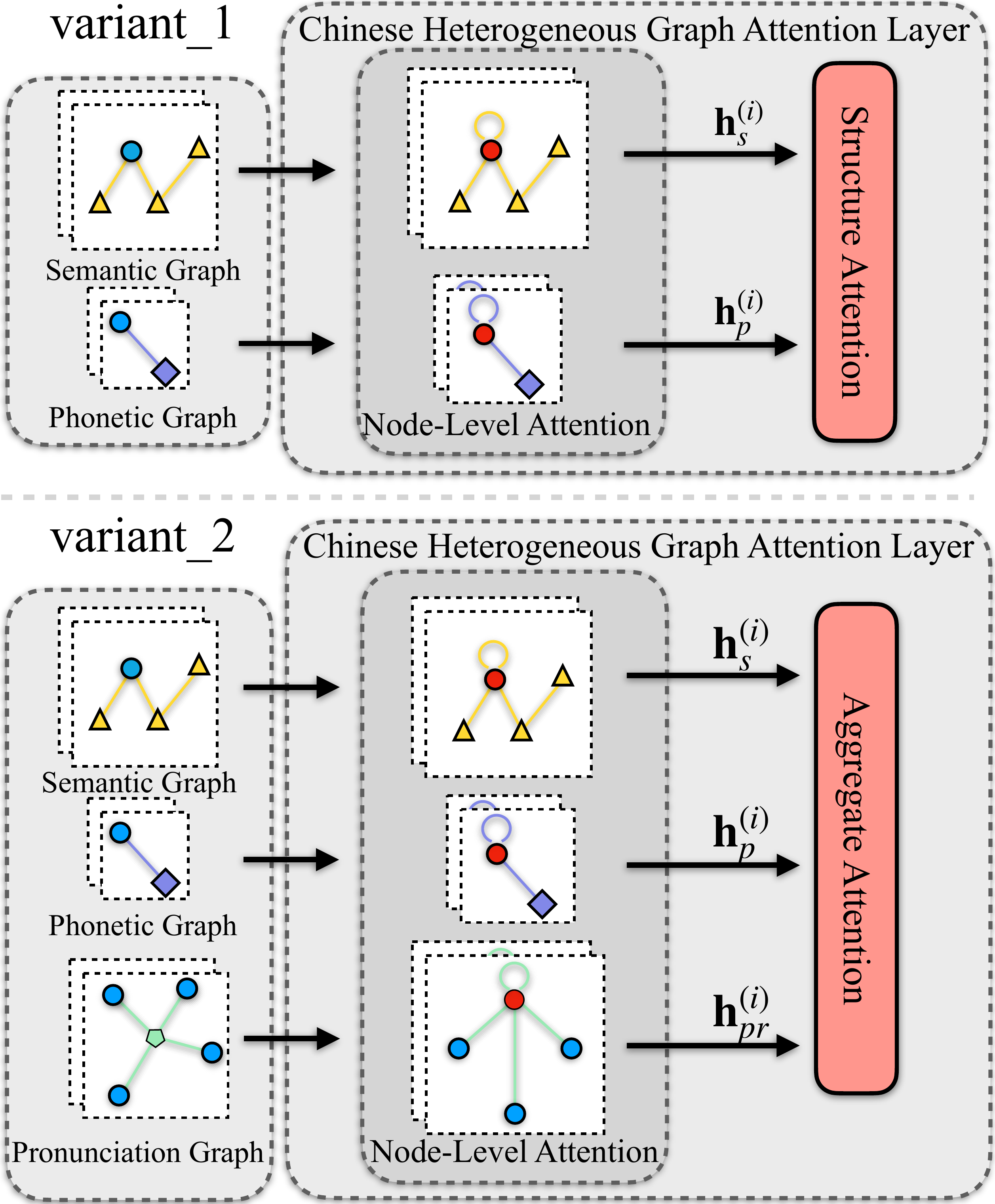}
\centering
\caption{Illustration of CHGAT layer's variants.}
\label{fig:variants}
\end{figure}

To validate the pronunciation node type included in our heterogeneous graph and the design consideration of three-level attention network, we build two variants (variant\_1 and variant\_2) of our network and compare them with our original model and FGAT. variant\_1 is obtained by removing the pronunciation node type and \textit{character-pronunciation-character} meta-path graph from the CHGAT layer in the original model (details in the upper part of Figure~\ref{fig:variants}). The aggregation attention layer used for aggregating pronunciation and structural information is no longer needed and therefore removed. In variant\_2 (shown in the lower part of Figure~\ref{fig:variants}), we remove the structure attention layer and use the aggregate attention layer directly.


On the \textbf{9,800 Names} dataset (split in 8:1:1 for training, validation, and test sets), the two variants show decreased performance from the original model but still remain more effective than the FGAT model as shown in Table~\ref{fig:variants}. The variant\_1 achieves a relative improvement of 1.27\% compared to FGAT, which suggests that the heterogeneous graph in variant\_1 obviously captures more structural information than the homogeneous one in FGAT.

Unsurprisingly, our model achieves a relative improvement of 1.26\% compared to variant\_2, which indicates that the multi-level attention network in the original model is more effective than the single-level attention network.

Though variant\_2 includes pronunciation information, it has a worse performance compared with variant\_1. This suggests that variant\_2's single-level attention does a very bad job in incorporating pronunciation information, such that it even introduces noises that undermine its performance. 

\begin{table}[htb]

\centering
\small
\begin{tabular}{c|c|c|c|c}

\hline
 & FGAT& variant\_1 & variant\_2 & our model \\
\hline
accuracy& 0.8010 & 0.8112  & 0.8071 & 0.8173 \\
\hline
\end{tabular}

\caption{Accuracy of FGAT, variant\_1, variant\_2, and our model trained and tested on \textbf{9,800 Names}.}
\label{tab:result3}
\end{table}

\subsection{Complexity Analysis}
We analyze the complexity of the baselines and our model. For PBERT, the complexity is $O(d n^{2})$, where $d$ denotes the feature dimension and $n$ is the sequence length (number of Pinyin letters in a name). FGAT has a word graph learning part that increases the complexity to $O(d n^{2} + L|V|d^{2} + L|E|d)$, where $L$ is the number of GNN layers, $|V|$ is the number of nodes and $|E|$ is the number of edges. Compared with FGAT, our model performs additional attention aggregations, adding $adp^{2}$ to the complexity, where $p$ denotes the number of meta-paths and $a$ is the number of aggregations. This is a slight increase since the complexity is largely determined by $d^{2}$. Besides, as the training happens offline and we do not need instant responses when guessing the genders, the complexity is acceptable in practice.

\section{Conclusion}
To address the lack of high-quality Chinese name-gender prediction tools and to facilitate the gender bias research for the underrepresented, we propose a heterogeneous graph attention model incorporating structural and pronunciation information of Chinese characters for Chinese name-gender prediction that outperforms all SOTA models. Besides, we open source a large-scale Chinese name-gender dataset as well as our source code. As a future step, we plan to extend our method to the many other tasks that involve Chinese character representations.

\section{Acknowledgments}
This project is partially supported by the Key Projects of Shanghai Soft Science Research Program, from the Science and Technology Commission of Shanghai Municipality (No. 22692112900).

 \bibliography{aaai23}
\end{document}